\pdfoutput=1

\documentclass[11pt]{article}

\usepackage{emnlp2021}

\usepackage{times}
\usepackage{latexsym}

\usepackage{breqn}
\usepackage{amsmath}
\usepackage{url}

\usepackage[T1]{fontenc}

\usepackage[utf8]{inputenc}

\usepackage{microtype}
\usepackage{comment}

\title{cushLEPOR: customising hLEPOR metric using Optuna for higher agreement with human judgments or pre-trained language model LaBSE}

\author{
          Lifeng Han$^1$,  Irina Sorokina$^{2}$, Gleb Erofeev$^{2}$,   \and  Serge Gladkoff$^2$ \\
         $^1$ ADAPT Research Centre, DCU, Ireland  \\ 
         $^2$ Logrus Global,  Translation \& Localization 
         \\ {\tt lifeng.han@adaptcentre.ie} 
         \\
         {\tt 
         gleberof, irina.sorokina, serge.gladkoff@logrusglobal.com}         }
\begin{document}
\maketitle{}

\begin{abstract}
Human evaluation has always been expensive while researchers struggle to trust the automatic metrics.
To address this, we propose to customise traditional metrics by taking 
advantages of the pre-trained language models (PLMs) and the limited available human labelled scores.
We first re-introduce the hLEPOR metric factors, followed by the Python  version we developed (ported) which achieved the automatic tuning of the weighting parameters in hLEPOR metric. 
Then we present the customised hLEPOR (cushLEPOR) which uses Optuna hyper-parameter optimisation framework to fine-tune hLEPOR weighting parameters towards better agreement to pre-trained language models (using LaBSE) regarding the exact MT language pairs that cushLEPOR is deployed to. 
We also optimise cushLEPOR towards professional human evaluation data based on MQM and pSQM framework on English-German and Chinese-English language pairs. 
The experimental investigations show cushLEPOR boosts hLEPOR performances towards better agreements to PLMs like LaBSE with much lower cost, and better agreements to human evaluations including MQM and pSQM scores, and yields much better performances than BLEU (data available at \url{https://github.com/poethan/cushLEPOR}). Official results show that our submissions win three language pairs including \textbf{English-German} and \textbf{Chinese-English} on \textit{News} domain via cushLEPOR(LM) and \textbf{English-Russian} on \textit{TED} domain via hLEPOR.

\end{abstract}

\section{Introduction}

Machine Translation (MT) is a rapidly developing research field that plays an important role in NLP area. MT started from 1950s as one of the earliest artificial intelligence (AI) research topics and gained a large improvement in the output quality in large resourced language pairs after the introduction of Neural MT (NMT) in recent years \cite{kalchbrenner13emnlp, cho2014encoder-decoder, DBLP:journals/corr/BahdanauCB14}. However, the challenge still remains in achieving human parity of MT output \cite{han-etal-2021-chinese}. Thus MT evaluation (MTE) continues to play an important role in aiding MT development from the aspects of timely and high quality evaluations, as well as reflecting the translation errors that MT systems can take advantages of for further improvement \cite{han-etal-2021-TQA}.
On one hand, human evaluations have long been criticised as expensive and unrepeatable. Furthermore, the inter- and intra-agreement levels from Human raters may  struggle to achieve a consistent and reliable score, unless done in rigour with highly trained and skilled evaluators \cite{Alekseeva-etal-2021-MonteCarlo}. 
On the other hand, even though researchers have claimed that the automatic evaluation metrics have reached much better performances in the category of system level evaluations of MT outputs, with high correlation to human judgements, the segment level performance is still a large gap from human experts' expectation \cite{google2021human_evaluation_TQA,barrault-etal-wmt2019-findings,barrault-etal-wmt2020_findings,han-etal-2013wmt-description,machacek-bojar-wmt2013-results}.

In the meantime, many pre-trained language models have been proposed and developed in very recent years and showing big advantages in different NLP tasks, for instance, BERT \cite{devlin-etal-2019-bert} and its further developed variants \cite{labse2020lm}. 
In this work, we take the advantages of both high performing automatic metric and pre-trained language model, aiming at one step further towards higher quality performing automatic MT evaluation metric from both system level and segment level perspectives. 

Among the evaluation metrics developed recent years, hLEPOR \cite{han2013hLEPOR_MTsummit,han-etal-2013wmt-description} is an augmented metric that include many evaluation factors with tunable weights assigned including precision, recall, word order (via position difference factor), and sentence length. It has also been applied by  researchers from different NLP fields including 
 natural language generation (NLG) \cite{NLG_evaluation2017emnlp_Novikova_etal,arxiv2021NLG}, natural language understanding (NLU) \cite{2021NLU_ruder_multilingual_evaluation},  automatic text summarization (ATS) \cite{bhandari-etal-2020-evaluating_summarization}, and searching \cite{search_eval2021Liu}, in addition to MT evaluation \cite{Marzouk2021CNL4mt}.

However, original hLEPOR has disadvantage of the manual tuning of its parameter weights which take a lot of human efforts.
We choose hLEPOR \cite{han2013hLEPOR_MTsummit,han-etal-2013wmt-description} as our baseline model, and use the very recent language model LaBSE \cite{labse2020lm} to achieve automatic tuning of its parameters thus aiming at reducing the evaluation cost and further boosting the performance. This system description paper is based on our earlier work, especially the training models \cite{cushLEPOR21MTsummit}.

The rest of the paper is organised as below: Section \ref{revisit_hLEPOR} revisits hLEPOR metric, its factors, advantages and disadvantages, Section \ref{proposed_model} introduces our Python ported version of hLEPOR and the further customised hLEPOR (cushLEPOR) using language models, Section \ref{experimental_eval} presents our experimental development and evaluation that we carried out on cushLEPOR metric using WMT historical data, Section \ref{to_wmt21} reserves space for our submission to this year WMT21 metrics task, and Section \ref{discussion} finishes this paper with discussions of our findings and possible future work. 


\section{Revisiting hLEPOR}
\label{revisit_hLEPOR}

hLEPOR is a further developed variant of LEPOR \cite{han2012lepor} metric which was firstly proposed in 2013 including all evaluation factors from LEPOR but using harmonic mean for grouping factors to produce final calculation score \cite{han2013hLEPOR_MTsummit}.
Its submission to WMT2013 metrics task achieved system level  highest  average correlating scores to human judgement on English-to-other (French, Spanish, Russian, German, Czech) language pairs by Pearson correlation coefficient (0.854) \cite{han2014lepor,machacek-bojar-wmt2013-results}. 
Other MT researchers also analysed LEPOR metric variant as one of the best performing segment level metric that was not significantly outperformed by other metrics using WMT shared task data \cite{DBLP:conf/naacl/GrahamBM15}.
hLEPOR is calculated by:

\begin{gather*} 
\text{\textit{h}LEPOR} = {Harmonic(w_{LP}LP},
\\ w_{NPosPenal}NPosPenal, w_{HPR}HPR)
\end{gather*}



\noindent
where \textit{LP} is a sentence length penalty factor which was extended from brevity penalty utilised in BLEU metric, \textit{NPosPenal} is for n-gram position difference penalty which captures the word order information, as bellow, where $MatchN_{hyp}$ and $MatchN_{ref}$ indicate the position number of matched words in hypothesis and reference sentences:

\begin{gather*} 
NPosPenal = e^{-NPD}
\\ NPD = \frac{1}{Length_{hyp}}\sum_{i=1}^{Length_{hyp}}|PD_i|
\\ |PD_i|= |MatchN_{hyp}-MatchN_{ref}|
\end{gather*}

\noindent The factor \textit{HPR} is the harmonic mean of Precision and Recall values. 

\begin{gather*} 
HPR = \frac{(\alpha+\beta)Precision x Recall}{\alpha Precision + \beta Recall}
\\ Precision = \frac{Aligned_{num}}{Length_{hypothesis}}
\\ Recall = \frac{Aligned_{num}}{Length_{reference}}
\end{gather*}

\noindent We refer the work \cite{han2014lepor,han-etal-2013wmt-description,han2012lepor} for detailed factor calculation with examples there.

The basic version of hLEPOR carries out similarity calculation between MT system outputs and reference translations, in the same language setting, based on the \textit{word surface level}  tokens. The hybrid hLEPOR metric also carries out similarity calculation based on POS sequences from system-output and reference text. To do this, POS tagging is needed as the first step, then hLEPOR(POS) calculation uses the same algorithms used for the word level similarity score hLEPOR(word). Finally, hybrid hLEPOR is a combination of both word level and POS level score. 
In this system submission work, with the time limitations, to make an easier to use customised hLEPOR, we take the basic version of hLEPOR, i.e. the word level similarity calculation and leave the hybrid hLEPOR into the future work.

The weighting parameters for the three main factors in original hLEPOR metric, i.e the ($w_{LP}$, $w_{NPosPenal}$, $w_{HPR}$) set, in addition to the other parameters inside each factor, were tuned by manual work based on development data (see \textit{Appendix} 
for detailed parameter value sets on each language pair for word-surface level evaluation, en $\Leftrightarrow$ cs/fr/de/es/ru). This is very time consuming, tedious, and costly.
In this work, we will introduce an automated tuning model for hLEPOR to customise it regarding deployed language pairs, which we name as cushLEPOR.


\section{Proposed Model}
\label{proposed_model}

\subsection{Python port of hLEPOR}
Original hLEPOR was published as Perl code \footnote{\url{https://github.com/poethan/LEPOR}}, in a non-portable format, which is not very suitable for modern AI/NLP applications, since they are using almost exclusively Python. Python is a programming language of choice for AI and machine learning (ML) tasks, thanks to its amazing ecosystem of open source or simply free libraries available to researchers and developers. However, hLEPOR was not available in NLTK \cite{nltk2009} or any other public Python libraries.
We therefore took original published Perl code and ported it to Python, carefully comparing the logic of original paper and the Perl implementation.
During this work we 
run both Perl code to reproduce the results of original code, and the new Python implementation.  
This work helped us to spot and fix at least three minor errors which did not significantly affected the score, but nevertheless we fixed the bugs of the Perl code.

While doing the porting we did also notice that hLEPOR parameter values were taken empirically and never explained in detail except for the suggested parameter setting table in the paper \cite{han2013hLEPOR_MTsummit,han-etal-2013wmt-description} for eight language pairs that were tested for the WMT2013 shared task, including EN-CZ/DE/FR/ES and the opposite direction. They were:

\begin{itemize}
    \item \textbf{alpha}: the tunable weight for recall
    \item \textbf{beta}: the tunable weight for precision
    \item \textbf{n}: words count before and after matched word in npd calculation
    \item \textbf{weight_elp}: tunable weight of enhanced length penalty
    \item \textbf{weight_pos}: tunable weight of n-gram position difference penalty
    \item \textbf{weight_pr}: tunable weight of harmonic mean of precision and recall 
\end{itemize}

The parameter values for hLEPOR as published in the publicly available Perl code were mannually tuned for  \textit{English-to-Czech/Russian (EN=>CS/RU) language pair} setting \cite{han2013hLEPOR_MTsummit,han-etal-2013wmt-description} as below:

\begin{itemize}
    \item \textbf{alpha} = 9.0, 
    \item \textbf{beta} = 1.0,
    \item \textbf{n} = 2,
    \item \textbf{weight_elp} = 2.0,
    \item \textbf{weight_pos} = 1.0,
    \item \textbf{weight_pr} = 7.0.
\end{itemize}

We refer to our \textbf{Appendix} for the manually tuned parameters for other language pairs available in the paper by \cite{han-etal-2013wmt-description,han2013hLEPOR_MTsummit} including English=>French/German (EN=>FR/DE) and Czech/Spanish/French/German=>English (CS/ES/FR/DE=>EN) and Russian=>English (RU=>EN) which was set up using CS=>EN without extra manual tuning. We 
came to the conclusion that we need to check whether these parameters are optimal, and find out whether better set of values exist to improve agreement with human judgement.

Because the different characteristics of each language, and language families, the evaluation of MT outputs would emphasis on different factors. For instance, word order factor reflected by n-gram position different penalty in hLEPOR (NPosPenal), can be with higher or lower weight for strict order languages and loose/flexible word order languages.
Thus, we assumed that hLEPOR optimisation towards different languages will generate corresponding different set of parameter values. We call this step language-specific optimisation, and it will save much cost and time to achieve an automatic tuning process.
The Python ported hLEPOR is available at Pypi \url{https://pypi.org/project/hLepor/}.

\subsection{Customised hLEPOR: cushLEPOR}
With the recent development of pre-trained neural language models and their effective applications in different NLP tasks, including question answering, language inference and MT, it becomes a natural question that why do not we apply them in MT evaluation as well.

Very recent work from Google team verified that MQM (Multi-dimension quality metric) \cite{MQM2014} and SQM (Scalar Quality Metrics) \cite{google2021human_evaluation_TQA} have good agreement with each other when they were carried out both by professional translators. However, this does not correlate to Mechanical Turk based crowd-sourced human evaluation that was carried out by general researchers or untrained online workers with low professional linguistic skills. It also reflected that crowd-sourced evaluation tends to favour very literal translations instead of better translations with more diverse meaning equivalent lexical choices. 

To customise hLEPOR (cushLEPOR) towards optimised parameter setting 
for deployed language pairs, we choose 
Optuna open source hyper-parameter optimisation framework \cite{Optuna2019kdd} to automate hyper-parameter search for best agreement between cushLEPOR and human experts evaluation wherever such data-set is available. 

SQM \cite{google2021human_evaluation_TQA} borrows WMT shared task settings to collect segment-level scalar rating, but set the score scale from 0 to 6
instead of 0 to 100. Professional translator labelled scores using SQM is named as pSQM.

We aim at optimising cushLEPOR parameters to obtain best agreement with pSQM 
scores. However, in practical situation, human evaluations are not often feasible to obtain due to the  constrains from both time and financial aspects.

We therefore propose to carry out an alternative optimisation model, i.e. 
customising 
cushLEPOR parameters towards pre-trained large scale language models, e.g. LaBSE (Language Agnostic BERT Sentence Embedding) model similarity score.

LaBSE model is built on BERT (Bidirectional Encoder Representations from Transformers) architecture and trained on monolingual (for dictionaries) and bilingual training data. LaBSE training data is filtered and processed.
The resulting sentence embeddings achieve excellent performance on measures of sentence embedding quality such as the semantic textual similarity (STS) benchmark and sentence embedding based transfer learning \cite{labse2020lm}.

LaBSE linguistic similarity score finds matching translations very well. The disadvantages, however, are high demand for computational resources (with GPUs), intensive application coding with requirement for ML skills, and slow performance.

The design 
of using optimised hLEPOR (cushLEPOR) in lieu of LaBSE similarity aims at developing a simple, high-performing, easy to run and not computationally demanding script to achieve results similar to high-end LaBSE similarity score, and 
hopefully towards human judgement. 
The cushLEPOR parameters can be optimised for agreement with any type of scores, such as pSQM, MQM, and LaBSE, etc.

Regarding the optimisation stage using Optuna, 
the task is to find the extremum values of continuous (not discrete) surface in a 6-dimensional space of six cushLEPOR parameters. The values of parameter set change continuously which means there’s an infinite number of parameter values; however 
it is not a differentiable situation mathematically, and there are gaps. Generally we cannot presume that it is a smooth surface.
Before Optuna, computational tools used to deploy a discrete mesh in such cases by using discretization method, which was less computationally intense than full scale continuous search on all possible values of parameter sets.

Optuna framework is currently one of the best Tree-structured Parzen Estimator (TPE) model implementations, which kind of estimators 
 converges to optimal solution in 200-300 epochs, and the method can work with continuous (real) parameters \cite{Hyper_param_2011nips}.

\section{Experimental Evaluations}
\label{experimental_eval}

The training and development data we used regarding MQM scores and pSQM labels is from the recent work by Google Research team on investigating into human evaluations based on WMT2020 shared task \cite{google2021human_evaluation_TQA} (data available at \url{https://github.com/google/wmt-mqm-human-evaluation}).

We first focus on English-to-German language (EN-DE) pair, which includes MQM and pSQM labels, acquired from  10 submission of WMT 2020, then take ZH-EN data-set. We refer to the paper \cite{google2021human_evaluation_TQA} for detailed MT system names and offering institutions.


Firstly, a multi-parameter optimisation against LaBSE for EN-DE language pair gave the following values for cushLEPOR parameters:

\begin{itemize}
    \item \textbf{alpha} = 2.97, 
    \item \textbf{beta} = 1.97,
    \item \textbf{n} = 4,
    \item \textbf{weight_elp} = 1.0,
    \item \textbf{weight_pos} = 14.97,
    \item \textbf{weight_pr} = 2.2.
\end{itemize}

This set of values reflected very different weighting systems in comparison to the original hLEPOR metric. For instance, 1) cushLEPOR assigned recall and precision much closer weight (2.97 vs 1.97) in comparison to hLEPOR (9.0 vs 1.0), 2) cushLEPOR chose 4-gram in chunk matching instead of bi-gram used in hLEPOR, 3) cushLEPOR assigned NPosPenal (n-gram position difference penalty) factor a very heavy weight against other two factors LP (length penalty) and HPR (harmonic mean of precision and recall) by (\textbf{14.97} vs 1.0 and 2.2) in comparison to hLEPOR which emphasised the weight on HPR (1.0 vs 2.0 and \textbf{7.0}). 
From these points of view, cushLEPOR trained on EN-DE language pair indicates the importance of the larger window context consideration during word matching, as well as the word order information reflected by n-gram (n value) and novel factor NPosPenal introduced by hLEPOR respectively.

\begin{figure}[!t]
\centering
\includegraphics*[width=0.45\textwidth]{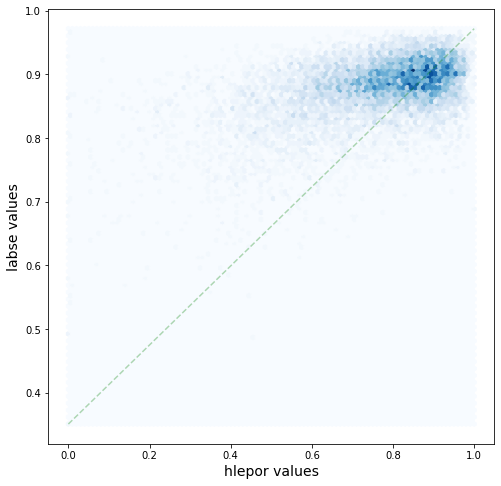}
\caption{Agreement with LaBSE: hLEPOR}
\label{fig:LaBSE_hLEPOR}
\end{figure}

\begin{figure}[!t]
\centering
\includegraphics*[width=0.45\textwidth]{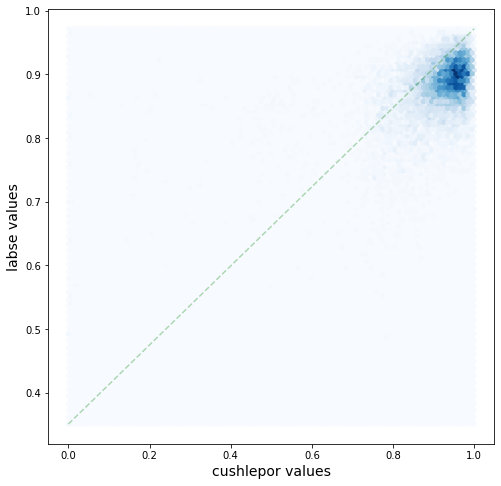}
\caption{Agreement with LaBSE: cushLEPOR}
\label{fig:LaBSE_cushLEPOR}
\end{figure}

This also reflected that LaBSE similarity is indeed a feasible goal for cushLEPOR optimisation.
The correlations of hLEPOR and cushLEPOR to LaBSE are shown in Fig. \ref{fig:LaBSE_hLEPOR} and \ref{fig:LaBSE_cushLEPOR}.

However, we found out that we were not able to decrease much on RMSE (Root Mean Square Error) score for cushLEPOR 
towards pSQM
, in comparison to original hLEPOR, (0.28 vs 0.29)
which does indicate that original hLEPOR empirically shows 
very good fit for pSQM type human evaluation, using the suggested parameter settings for EN-DE \cite{han-etal-2013wmt-description,han2013hLEPOR_MTsummit} as bellow.

\begin{itemize}
    \item \textbf{alpha} = 9.0, 
    \item \textbf{beta} = 1.0,
    \item \textbf{n} = 2,
    \item \textbf{weight_elp} = 3.0,
    \item \textbf{weight_pos} = 7.0,
    \item \textbf{weight_pr} = 1.0.
\end{itemize}

\begin{figure}[!t]
\centering
\includegraphics*[width=0.45\textwidth]{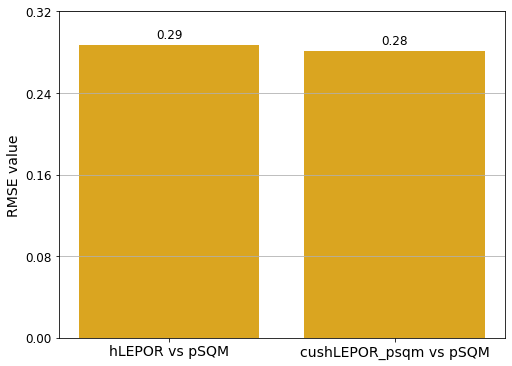}
\caption{RMSE: hLEPOR vs cushLEPOR to pSQM (lower score is better)}
\label{fig:RMSE_hLEPORs_pSQM}
\end{figure}

\begin{figure}[!t]
\centering
\includegraphics*[width=0.45\textwidth]{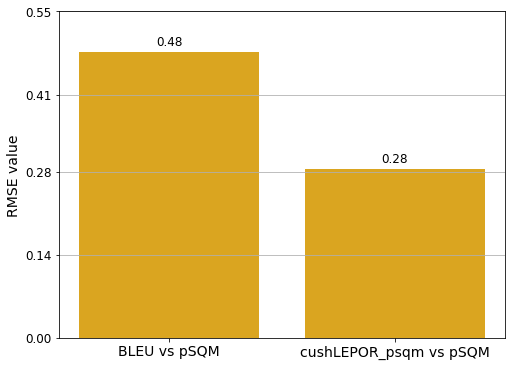}
\caption{RMSE: BLEU vs cushLEPOR to pSQM (lower score is better)}
\label{fig:RMSE_BLEU_cushLEPOR_pSQM}
\end{figure}

The RMSE value between pSQM and hLEPOR, vs pSQM and cushLEPOR is shown in Fig. \ref{fig:RMSE_hLEPORs_pSQM}. However, it indeed shows much better performance then BLEU metric, as in Fig. \ref{fig:RMSE_BLEU_cushLEPOR_pSQM} (0.28 vs 0.46).

\begin{figure}[!t]
\centering
\includegraphics*[width=0.45\textwidth]{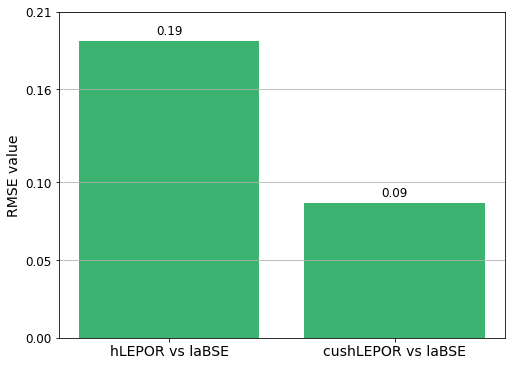}
\caption{RMSE: hLEPOR vs cushLEPOR to LaBSE (lower score is better)}
\label{fig:RMSE_hLEPORs_LaBSE}
\end{figure}

Optuna did optimise cushLEPOR against LaBSE very well, halving the RMSE distance between LaBSE and cushLEPOR as compared to original hLEPOR, shown in Fig. \ref{fig:RMSE_hLEPORs_LaBSE}.

\begin{figure}[!t]
\centering
\includegraphics*[height=2.1in,width=3in]{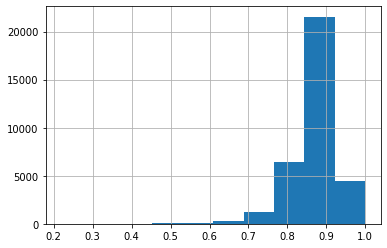}
\caption{Score Distribution: tune on LaBSE}
\label{fig:tune_LaBSE}
\end{figure}

\begin{figure}[!t]
\centering
\includegraphics*[width=0.45\textwidth]{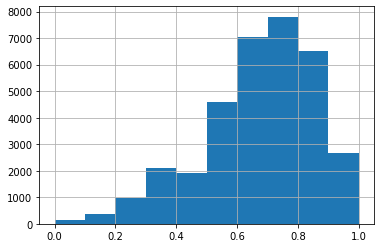}
\caption{Score Distribution: tune on pSQM}
\label{fig:tune_pSQM}
\end{figure}

The performances of tuning on LaBSE and pSQM are shown in Fig. \ref{fig:tune_LaBSE} and \ref{fig:tune_pSQM} respectively. The horizontal axis is the score value (0, 1) and the vertical axis is the sentence number that falls into the corresponding score intervals.

From the score distribution visualisation, it reflects the tuning on pSQM has a larger covered error types while LaBSE is less sensitive to some errors that human experts would spot out. As shown on these charts, pSQM human rating shows much wider "tail" of "low score ratings", while LaBSE rating is much more focused. 
The reason is that LaBSE similarity model underestimates the severity of errors and error types, while humans analyse the meaning and assign proper error penalties in more diverse setting. As an example, the sentence "The comet did not struck the Earth this time." and "The comet did struck the Earth this time." has very close lexical similarity, 
but the meaning is very different, in this case ``opposite''. LaBSE similarity score would not assign significant penalty to such difference, while human will treat it as a major error.
This difference plays a crucial role for reliable translation quality evaluation.

\section{Submission to WMT21}
\label{to_wmt21}
For WMT2021 Metrics Task, we submitted our cushLEPOR system scores for zh=>en and en=>de language pairs, both segment-level and system-level evaluation. The training and development set we used are exact the ones from last section (Section \ref{experimental_eval}). We can not tune our cushLEPOR model parameters on en=>ru language pair from the WMT21 official data, because the human labelled MQM and pSQM scores as validation data that cushLEPOR requires do not exist from last year WMT20 set. Instead, we submitted hLEPOR metric for EN=>RU using the parameter settings in hLEPOR as mentioned in the last section.
We carried out evaluation on all four official data-sets: \textbf{newstest2021} (traditional task), \textbf{florestest2021} (sentences translated as part of the WMT News translation task), \textbf{tedtalks} (additional sets of sentences translated by WMT21 translation systems in the TED talks domain), and \textbf{challengeset} (synthetic outputs generated specifically to challenge automatic metrics).



\subsection{Submitted Parameter Setting}
The optimised parameter values set for our 
zh=>en submission to WMT21 is displayed below.

For cushLEPOR(LM) using LaBSE training:
\begin{itemize}
    \item \textbf{alpha} = 2.85, 
    \item \textbf{beta} = 4.73,
    \item \textbf{n} = 1,
    \item \textbf{weight_elp} = 1.01,
    \item \textbf{weight_pos} = 11.13,
    \item \textbf{weight_pr} = 4.62
\end{itemize}

For cushLEPOR(pSQM) using  professional translator labelled SQM training:
\begin{itemize}
    \item \textbf{alpha} = 9.09, 
    \item \textbf{beta} = 3.55,
    \item \textbf{n} = 3,
    \item \textbf{weight_elp} = 1.01,
    \item \textbf{weight_pos} = 14.98,
    \item \textbf{weight_pr} = 1.57
\end{itemize}

The optimised parameter values set for our 
en=>de submission to WMT21 is displayed below:

For cushLEPOR(LM) using LaBSE training:
\begin{itemize}
    \item \textbf{alpha} = 2.95, 
    \item \textbf{beta} = 2.68,
    \item \textbf{n} = 2,
    \item \textbf{weight_elp} = 1.0,
    \item \textbf{weight_pos} = 11.79,
    \item \textbf{weight_pr} = 1.87
\end{itemize}

For cushLEPOR(pSQM) using  professional translator labelled SQM training:
\begin{itemize}
    \item \textbf{alpha} = 1.13, 
    \item \textbf{beta} = 1.71,
    \item \textbf{n} = 2,
    \item \textbf{weight_elp} = 1.06,
    \item \textbf{weight_pos} = 11.90,
    \item \textbf{weight_pr} = 1.01
\end{itemize}

\subsection{Official Results from Metrics Task}
The official results from WMT2021 Metrics task show that cushLEPOR(LM) ranks in the \textbf{first cluster} in performance on \textbf{News test} data with single reference evaluated on overall English-to-German, Chinese-to-English and English-to-Russian where professional human evaluation data is available (\textit{Ref.} Table 8 ``Metric rankings based on pairwise accuracy'' in Findings paper \cite{freitag-etal-2021metrics-findings}). 
Furthermore, in the language specific ranking, cushLEPOR(LM) also wins \textbf{English-to-German} and \textbf{Chinese-to-English} language pairs, including TED data condition. Our hLEPOR baseline metric wins \textbf{English-to-Russian} TED domain language specific ranking (Ref. Table 12 ``Summary of language-specific results'' in the official findings paper \cite{freitag-etal-2021metrics-findings}).
The official result on ``System-level Pearson correlations for \textbf{English-to-German}'' (Table 23 of findings) shows that cushLEPOR(LM) achieves score
0.938 in News domain, ranking \textit{number 1 in Cluster 1 metrics}, out of overall 29 metric submissions.

\section{Discussions and Future Work}
\label{discussion}

In this work, we described cushLEPOR, a customised hLEPOR metric 
which can be automatically trained and optimised using both human labelled MQM scores, as well as large scale pre-trained language model (LM) LaBSE towards better agreement to human experts level judgements and distilled LM performance respectively, and reducing cost at the meantime, e.g. the manual tuning from hLEPOR and high computational demand from LMs.

We also optimised cushLEPOR towards human translators' evaluation scores, i.e. pSQM, which showed much improved performance than BLEU and original hLEPOR (with default parameters).
Our research is in line with the MT evaluation guideline suggestions from the very recent work \cite{marie-etal-2021acl-scientific} that better evaluation metrics in correlation to human judgement shall be tested and deployed. Or human judgements shall be carried out directly wherever possible.

We have some findings during the experimental investigation: 1) cushLEPOR trained on LaBSE can replace LaBSE to carry out similarity calculation task in MT evaluation, which is much more light weighted and low cost from computational power and complexity point of view. 2) we can choose alternative pre-trained language models (LMs) in the future to boost performance. 3) this cushLEPOR optimisation framework proves to be functional,  offering high performance towards pre-trained LMs, much improved agreement of cushLEPOR to LaBSE scores in comparison to hLEPOR (
as in Figure \ref{fig:LaBSE_hLEPOR} and \ref{fig:LaBSE_cushLEPOR}). 4) optimised cushLEPOR achieves better agreement towards professional translator's evaluation (pSQM).

Optuna, the hyper parameter optimisation toolkit we used, can generate different set of cushLEPOR parameter values in different runs, which could be an consistency issue. However, we believe it optimises 
 the performance of cushLEPOR towards the highest agreement to the reference scoring (pre-trained LMs or human evaluations), but not to ensure the same set of parameter values to be generated,  so this will not be a problem. 
We will carry out further analysis on this aspect in the future work.

The hybrid version of hLEPOR \cite{han2013hLEPOR_MTsummit} use POS features to function as pseudo synonyms to capture alternative correct translations. However it relays on POS taggers for target language, which does not exist for newly proposed languages, and its tagging accuracy may be low, and it costs extra processing steps. In the future work, we plan to carry out integrated model which combine the POS tagging as a command function in data pre-processing for hybrid  cushLEPOR.

Overall, cushLEPOR achieved the first cluster performances in \textit{News} Domain data on \textit{Chinese-English} and \textit{English-German} in WMT2021 Metrics task, while hLEPOR wins \textit{TED} domain data on \textit{English-Russian} \cite{freitag-etal-2021metrics-findings}. In the future work, we plan to carry out optimisation of cushLEPOR on more language pairs as well as more domains. We will keep our updated parameter set for extended languages and domains available on our cushLEPOR open-sourced platform.

\section*{Acknowledgements}
The ADAPT Centre for Digital Content Technology is funded under the SFI Research Centres Programme (Grant 13/RC/2106) and is co-funded under the European Regional Development Fund.  We thank the following open-source teams: Google Research for sharing their human evaluation data on MQM and pSQM scores,  Optuna open-source hyper-parameter optimisation framework, and LaBSE pre-trained LMs.

\bibliography{anthology,custom}
\bibliographystyle{acl_natbib}

\section*{Appendices}
\label{appendix}
\section*{Appendix A: hLEPOR parameters}
The word level hLEPOR default parameters manually tuned for WMT2013 MT evaluation task across language pairs \cite{han-etal-2013wmt-description,han2013hLEPOR_MTsummit} are displayed as below. Both Python (\url{https://pypi.org/project/hLepor/}) and Perl (\url{https://github.com/lHan87/aaron-project-hlepor}) version codes can be applied to:

On English-to-Czech/Russian (EN=>CS/RU):

\begin{itemize}
    \item \textbf{alpha} = 9.0, 
    \item \textbf{beta} = 1.0,
    \item \textbf{n} = 2,
    \item \textbf{weight_elp} = 2.0,
    \item \textbf{weight_pos} = 1.0,
    \item \textbf{weight_pr} = 7.0.
\end{itemize}

On English-to-German (EN=>DE): 

\begin{itemize}
    \item \textbf{alpha} = 9.0, 
    \item \textbf{beta} = 1.0,
    \item \textbf{n} = 2,
    \item \textbf{weight_elp} = 3.0,
    \item \textbf{weight_pos} = 7.0,
    \item \textbf{weight_pr} = 1.0.
\end{itemize}

On Czech / Spanish / Russian to English (CS/ES/RU =>EN): 

\begin{itemize}
    \item \textbf{alpha} = 1.0
    \item \textbf{beta} = 9.0
    \item \textbf{n} = 2
    \item \textbf{weight_elp} = 2.0
    \item \textbf{weight_pos} = 1.0
    \item \textbf{weight_pr} = 7.0
\end{itemize}

On German/French-to-English (DE/FR=>EN) and English-to-Spanish/French (EN=>ES/FR): 

\begin{itemize}
    \item \textbf{alpha} = 9.0
    \item \textbf{beta} = 1.0
    \item \textbf{n} = 2
    \item \textbf{weight_elp} = 2.0
    \item \textbf{weight_pos} = 1.0
    \item \textbf{weight_pr} = 3.0
\end{itemize}

\end{document}